\documentclass[letterpaper]{article} % DO NOT CHANGE THIS
\usepackage{aaai2026}  % DO NOT CHANGE THIS
\usepackage{times}  % DO NOT CHANGE THIS
\usepackage{helvet}  % DO NOT CHANGE THIS
\usepackage{courier}  % DO NOT CHANGE THIS
\usepackage[hyphens]{url}  % DO NOT CHANGE THIS
\usepackage{graphicx} % DO NOT CHANGE THIS
\usepackage{natbib}  % DO NOT CHANGE THIS AND DO NOT ADD ANY OPTIONS TO IT
\usepackage{caption} % DO NOT CHANGE THIS AND DO NOT ADD ANY OPTIONS TO IT
\usepackage{algorithm}
\usepackage{algorithmic}
\usepackage{booktabs}
\usepackage{fontawesome}
\usepackage{pifont}
\usepackage{placeins}
\usepackage{newfloat}
\usepackage{listings}
\usepackage{soul}
\usepackage{xcolor}

\newcommand{\name}{TransLight}

\newcounter{checksubsection}
\newcounter{checkitem}[checksubsection]

\DeclareCaptionStyle{ruled}{labelfont=normalfont,labelsep=colon,strut=off} % DO NOT CHANGE THIS
\floatstyle{ruled}
\newfloat{listing}{tb}{lst}{}
\floatname{listing}{Listing}
\title{\name{}: Image-Guided Customized Lighting Control with Generative Decoupling}
\author{
    Zongming Li\textsuperscript{\rm 1}\textsuperscript{\rm 2}\footnotemark[1], Lianghui Zhu\textsuperscript{\rm 1}, Haocheng Shen\textsuperscript{\rm 2}, Longjin Ran\textsuperscript{\rm 2}, Wenyu Liu\textsuperscript{\rm 1}, Xinggang Wang\textsuperscript{\rm 1}\footnotemark[2]
}
\affiliations{
    \textsuperscript{\rm 1} School of EIC, Huazhong University of Science and Technology\\
    \textsuperscript{\rm 2} vivo Mobile Communication Co., Ltd
}

\usepackage{bibentry}
\begin{document}

\twocolumn[{
\renewcommand\twocolumn[1][]{#1}
\maketitle
\begin{center}
    \captionsetup{type=figure}
    \includegraphics[width=1\linewidth]{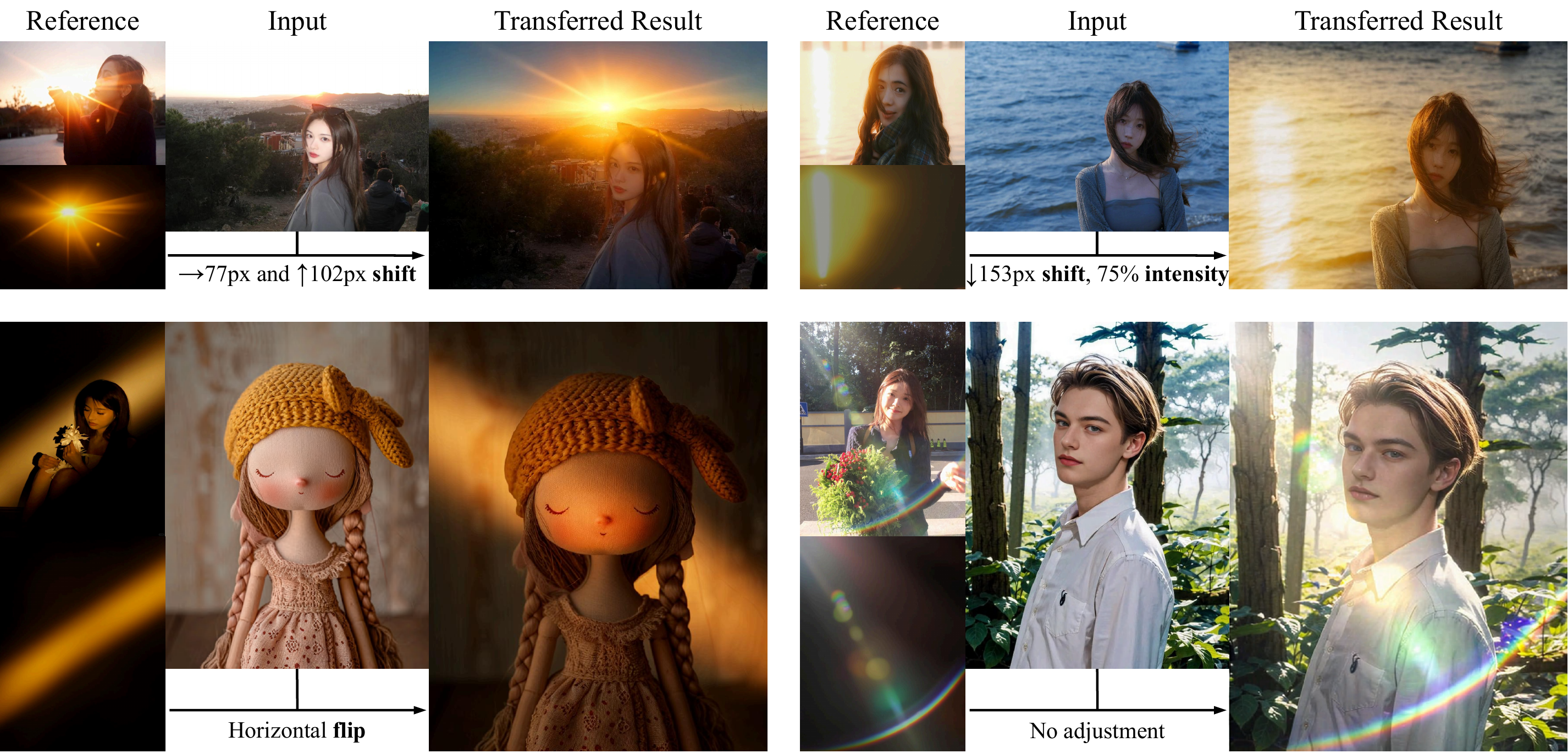}
    \caption{
    \textbf{Transfer light effects from reference to target image.}
    To the best of our knowledge, the proposed \name{} is the first method that high-fidelity transfers light effects from reference images to target images.
    We first extract the light effect, as shown below the reference image, and then composite it onto the user input image.
    Our \name{} allows flexible control over the position and direction of the light effects, thereby yielding visually authentic results with a high degree of freedom.
    }
    \label{fig:first_page_vis}
\end{center}
}]

\begin{abstract}
\textcolor{black}{Most existing illumination-editing approaches fail to simultaneously provide customized control of light effects and preserve content integrity. This makes them less effective for practical lighting stylization requirements, especially in the challenging task of transferring complex light effects from a reference image to a user-specified target image.}
To address this problem, we propose \textbf{\name{}}, a novel framework that enables high-fidelity \textcolor{black}{and high-freedom} transfer of light effects.
Extracting the light effect from the reference image is the most critical and challenging step in our method.
The difficulty lies in the complex geometric structure features embedded in light effects that are highly coupled with content in real-world scenarios.
To achieve this, we first present Generative Decoupling, where two fine-tuned diffusion models are used to accurately separate image content and light effects, generating a newly curated, million-scale dataset of image–content–light triplets.
Then, we employ IC-Light as the generative model and train our model with our triplets, injecting the reference lighting image as an additional conditioning signal.The resulting \name{} model enables customized and natural transfer of diverse light effects.
Notably, by thoroughly disentangling light effects from reference images, our generative decoupling strategy endows \name{} with highly flexible illumination control.Experimental results establish \name{} as the first method to successfully transfer light effects across disparate images, delivering more customized illumination control than existing techniques and charting new directions for research in illumination harmonization and editing.
\end{abstract}

\renewcommand{\thefootnote}{\fnsymbol{footnote}} 
\setcounter{footnote}{0} % 
\footnotetext{* Work was done during Zongming Li's internship at vivo Mobile Communication Co., Ltd.}
\footnotetext{ \dag  \   Corresponding author (\url{xgwang@hust.edu.cn})}

\section{Introduction}

Recent rapid iteration of image illumination editing methods~\citep{NLT, everlight, scribblelight, spotlight, switchlight, relightful, diffusionlight, dilightnet, neural, ic_light, lightlab, synthlight} has led to significant advancements in manipulating lighting attributes within the field of image editing. Relighting of human portraits, as a specific branch in this area, holds broad application prospects and high practical value. 
\textcolor{black}{Existing human portraits relighting approaches can generally be divided into two categories: visual-guided and text-guided illumination adjustment. Visual-guided methods~\citep{switchlight, relightful, synthlight} often employ high dynamic range (HDR) maps or additional background images to composite human portraits into new lighting environments. However, this typically leads to a loss of content integrity in the original image. Text-guided methods~\citep{ic_light, text2relight} accept textual descriptions of the desired light effects as input to perform image relighting. It is evident that textual descriptions alone are insufficient to achieve customized control over lighting properties such as direction and position.}
Although these methods can achieve striking visual effects, the application scenarios for relighting human portraits through background replacement or text prompts remain limited.
In real-world image editing scenarios, users may prefer personalized lighting modifications (e.g., adding or removing specific light effects) while preserving the original content of the image. 
For example, a user might request: \textit{``Add a rainbow lens flare similar to the reference image to my outdoor travel photo."}
Existing relighting approaches often struggle to fulfill such customized lighting stylization requirements.

In this work, we focus on the highly challenging and newly defined task of transferring customized light effects from a reference image to a target image. To achieve this, we first need to address two main challenges: (1) \textit{decoupling light effects from content to prevent content leakage} and (2) \textit{constructing a suitable data structure for light effect transfer training.}
Although this task is broadly similar to well-established style transfer, both aiming to replicate specific attributes from one reference image onto another image, it requires additional consideration of geometric information such as the direction, structure, and position of the light effects in the reference image. This leads to a stronger coupling between the light effects and the content, making many existing strategies~\citep{instantstyle, instantstyle-plus, csgo, swapping-attention} designed to mitigate content leakage in style transfer task less applicable to our case. 
Aiming at the challenges mentioned above, we propose \textbf{Generative Decoupling} as the solution.
Specifically, we fine-tune two diffusion models using natural images without light effects and light material images, enabling them to remove and extract light effects from input images.
After sufficient training, our light extraction model is capable of effectively isolating the most prominent light effects without incorporating any other content. This largely avoids the issue of content leakage in most cases.
Moreover, we construct over one million image-content-light triplets for light effect transfer task training using our generative decoupling strategy.

After that, we propose \name{}, the first method capable of transferring light effects from a reference image to another image.
We employ IC-Light as the fixed generative network and train our model using the image-content-light triplets.
During inference, the target light effect is decoupled from the reference image via generative decoupling, effectively mitigating content leakage and enabling greater transfer flexibility.
Users can freely adjust the transferred result by modifying the position and orientation of the light effect image.
Experimental results demonstrate that our \name{} possesses a powerful capability for customized light effects transfer. It can retain the background information of the source images while naturally incorporating target light effects from reference images as shown in Figure~\ref{fig:first_page_vis}. 
Moreover, our method achieves a Light FID score of 6.02, outperforming the SOTA illumination editing method IC-Light, which obtains 10.05.

The main contribution can be summarized as follows:
\begin{itemize}
    \item
    We propose \textbf{Generative Decoupling} to decouple content and light in natural images using two diffusion models. 
    Among them, light removal model eliminates prominent light effects from natural images, \textit{e.g.}, lens flares, Tyndall effects, \textit{etc}.
    Light extraction model is the first capable of separating light effects from images without retaining any original content.
    \item We construct an image-content-light triplet generation pipeline based on our generative decoupling strategy. We first select images with prominent light effects from a massive database, and then use our light removal and light extraction models to generate the corresponding content and light effects images, respectively. After filtering step, we obtain over one million training samples.
    \item We introduce \textbf{\name{}}, the first image-guided method that allows accurate transfer of light effects from a reference image to a target image. % 
    \textcolor{black}{
    Our method supports translation, flipping, and numerical scaling of the extracted light effects, enabling adjustments to the position, orientation, and intensity of the light effects in transferred result.
    Experimental results show that our \name{} achieves visually striking light effects across a wide range of scenarios.
    }
\end{itemize}

\section{Related Work}
\textcolor{black}{\subsection{Image Style Transfer}
Image style transfer aims to transfer the style of a reference image to the target content image~\citep{styleID, styleshot, stylediffusion, inversion, b-lora, instantstyle-plus, deadiff, stylestudio, csgo}.
This process is fundamentally similar to light effects transfer, which aims to replicate the intrinsic attributes of a reference image onto another image, while preserving the content consistency of the target image.
In this task, a well-studied problem is how to disentangle the content and style of the reference image to avoid content leakage.
StyelAligned~\citep{stylealigned} proposes replacing full attention with shared attention during style injection to reduce content leakage.
FreeTuner~\citep{freetuner} employs a disentanglement strategy that separates the generation process into two stages to effectively mitigate concept entanglement.
InstantStyle~\citep{instantstyle} and Swapping self-attention~\citep{swapping-attention} minimize the introduction of content information by injecting style features into specific attention layers of the generation network.
The aforementioned methods are specifically designed to disentangle abstract stylistic attributes from images and have proven effective in mitigating content leakage. However, they lack the capability to precisely perceive the geometric and structural characteristics of light effects, making it difficult to extract light. As a result, these approaches are not well-suited for the task of light effects transfer.
}

\subsection{Light Removal} Light removal aims to eliminate unwanted illumination from images while meticulously preserving the underlying scene content. The most representative application is the removal of flares from images~\citep{nighttime, removing, unsupervised, SIFR}. ~\citep{flare_removal} collected 2K real reflective flare data using a camera to train the first general lens flare removal network. After that, ~\citep{flare7k} further constructed 7K synthetic flare data to reduce the domain gap with real-world scenarios.
~\citep{SIFR} propose the first unpaired flare removal dataset and present a deep framework with light source aware guidance for single-image flare removal.
Recently, Illuminet~\citep{illuminet} proposed a two-stage pixel-to-pixel generative model based on the U-Net~\citep{u-net} model, that achieved both image flare removal and image light enhancement.
In addition, LuminaBrush~\citep{luminabrush} convert images to ``uniformly-lit" appearances by fine-tuning the DiT~\citep{dit} model through LoRA~\citep{lora}.
While these methods work well in specific situations, their limited training dataset makes it hard for them to perform satisfactorily in complex scenes, which means they can not adequately support the our decoupling requirements.

\subsection{Lighting Control with Diffusion Models}
Diffusion models have recently found widespread application in highly challenging tasks that demand precise control over image illumination, including portrait relighting and indoor lighting adjustment.
For portrait relighting, methods such as DiFaReli~\citep{difareli}, Lite2Relight~\citep{lite2relight}, and Holo-Relighting~\citep{holo} specifically target the relighting of human faces.
Relightful Harmonization~\citep{relightful} and Total Relighting~\citep{total_relighting} manipulate the illumination of the foreground portrait using background conditions.
SwitchLight~\citep{switchlight} controls portrait illumination by using high dynamic range (HDR) images as conditions.
Text2Relight~\citep{text2relight} generates relighted portrait using text prompt while keeping the original contents. Furthermore, IC-Light~\citep{ic_light}, trained extensively on large-scale datasets using the principle of light transport consistency, achieves impressive capabilities in text prompt and background controllable portrait illumination generation.
For indoor lighting adjustment, common operations include introducing new light sources or regulating the illumination intensity of those already present in a scene~\citep{latent, stylelight, lightlab}.
SpotLight~\citep{spotlight} control the local lighting of an object through a coarse shadow. 
By training a ControlNet~\citep{controlnet}, ScribbleLight~\citep{scribblelight} controls image light effects based on user scribbles.
While the aforementioned light control methods achieve good results in specific scenarios, their customized control over light effects is highly limited. They cannot precisely and comprehensively adjust details such as light position, shape, and type while preserving image content. Our \name{} aims to provide a solution.
\section{Method}

\begin{figure*}[ht]
\centering
\includegraphics[width=0.97\linewidth]{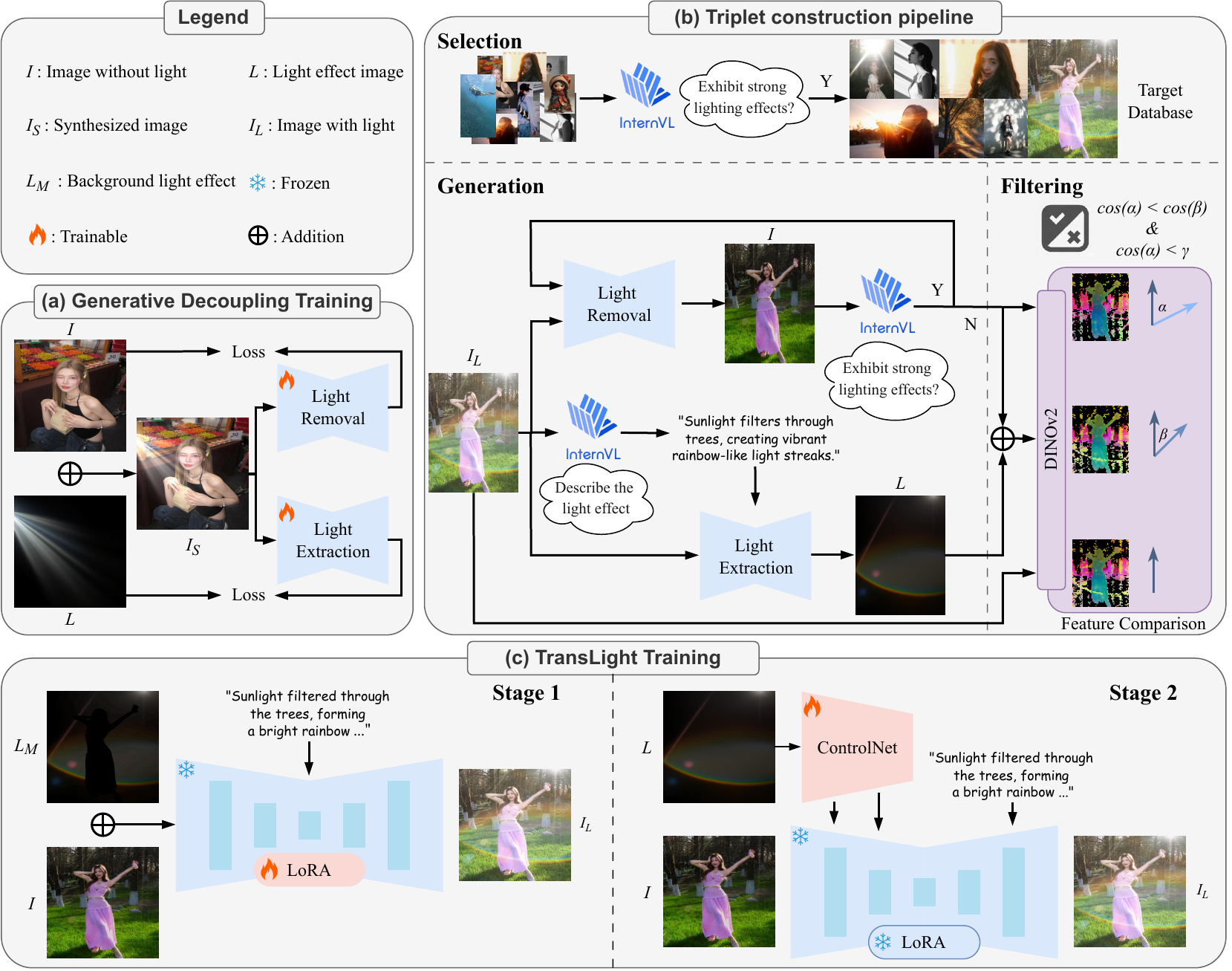}
\caption{\textcolor{black}{\textbf{Overall framwork.} (a) We fine-tune two diffusion models initialized with the weights of IC-Light. During training, the input synthesized image $I_S$ is the direct addition result of no light image $L$ and light material image $L$.
(b) This pipeline involves selecting relevant data using a vision language model, decoupling image content and lighting via light removal and extraction models, and filtering out poor generation results.
(c) Our \name{} training consists of two stages: LoRA fine-tuning for content-preserving lighting editing and ControlNet training to inject light effects.
}
}
\label{fig:whole_system}
\end{figure*}

\textcolor{black}{We systematically illustrate the overall framework for achieving the task of light effect transfer in Figure~\ref{fig:whole_system}. This framework comprises three key modules: the model training of generative decoupling, the image-content-light triplet construction pipeline, and the training of our \name{}.}

\subsection{Generative Decoupling}
\label{sec:generative_decoupling}
We use two diffusion models to decouple the image content and light effects, which we refer to as generative decoupling.
We design a simple yet effective strategy to train our light removal and light extraction models.
Figure~\ref{fig:whole_system} (a) shows our training strategy.
We first employ InternVL2.5~\citep{internvl} to select 1M images without light effects from a massive database. Then, we collect 100K light material images, all featuring pure black backgrounds and containing no objects other than light effects.
Specifically, we generate approximately 90K light material images using FLUX.1-schnell~\citep{flux} and collect an additional 10K real images from public datasets~\citep{flare7k, flare_removal}. After that, we fine-tune IC-Light~\citep{ic_light} directly, effectively leveraging its lighting adjustment capabilities acquired from training on large-scale in-the-wild images.
The synthesis of no light image $I$ and light material $L$ into $I_S$ involves no complex operations and can be expressed by the following formula:
\begin{equation}
\label{eq:synthesis}
    I_S = a I + b L,
\end{equation}
where $a, b \sim U(0, 1)$. Experimental results demonstrate that even with such simple synthesis operations, our fully trained generative decoupling model achieves excellent performance in real-world scenarios.

\subsection{Image-content-light Triplet Construction Pipeline}
\label{sec:dataset_pipeline}

To make light effects transfer possible, we construct a large volume of image-content-light triplets.
As shown in Figure~\ref{fig:whole_system} (b), our triplet construction pipeline contains three steps.

\textbf{Selection.} Initially, we select target data from a massive proprietary database. We employ InternVL2.5 for this large-scale batch selection.
We use highly stringent prompts to instruct the vision language model to determine if images contain strong light effects. 

\textbf{Generation.} We use light removal model and light extraction model to decompose an image with strong light effects into an image containing only content and an image containing only light effects. For the light removal process, we use InternVL2.5 to evaluate the generated results. If the results do not meet requirements, we perform a second round of light removal. In the light extraction process, we also use InternVL2.5 to generate a description of the image's light effects. This description, when combined with the image as input to the light extraction model, notably enhances the model's ability to separate light.

\textbf{Filtering.} We perform similarity-based filtering on the generated results to improve training data quality, discarding samples that exhibit notably poor generation outcomes. Specifically, after decoupling the input image $I_L$ into a no light image $I$ and a light effect image $L$, we obtain $I_S$ using the simple synthesis strategy from Equation~\ref{eq:synthesis}, with $a$ and $b$ fixed at 1. Then, we extract features from $I_S$, $I_L$, and $I$ separately using the DINOv2~\citep{dinov2}, and compute the cosine similarity of these features. The specific calculation method is as follows:
\begin{equation}
\label{eq:cos_alpha}
    cos(\alpha) = Sim_{cos}(\mathcal{D}(I_L), \mathcal{D}(I)),
\end{equation}
\begin{equation}
\label{eq:cos_beta}
    cos(\beta) = Sim_{cos}(\mathcal{D}(I_L), \mathcal{D}(I_S)),
\end{equation}
where $\mathcal{D}$ stands for DINOv2 and $Sim_{cos}$ stands for cosine similarity.
During our triplet construction, we aim for a significant degree of light effect removal, which necessitates that the similarity between $I$ and $I_L$ falls below a threshold $\gamma$. Concurrently, we desire a salient and clean separated light effect, thus expecting an increased similarity between $I_S$ and $I_L$. This means that the feature similarity among these images should satisfy the following expression:
\begin{equation}
\label{eq:judge}
    (cos(\beta) > cos(\alpha)) \land
    (cos(\alpha) < \gamma),
\end{equation}
where $\gamma$ is set to 0.98.

\subsection{\name{}}
\label{sec:\name{}}

Figure~\ref{fig:whole_system} (c) shows the training process of our \name{}.

\textcolor{black}{\textbf{Training stage 1.}}
First, we use the SD1.5~\citep{sd} version of the IC-Light~\citep{ic_light} as our generation network. This is done to fully leverage its powerful illumination editing capabilities, which were acquired through training on large-scale datasets. We freeze the parameters of the generation network during training. We first perform LoRA~\citep{lora} fine-tuning on the generation network, with the goal of enabling the model to preserve the background unchanged during the generation process and enhance its content consistency. 
\textcolor{black}{The input during this stage is a simple addition of the content image and the background light effects image.
This stage aims to enable the model to generate realistic images from composite inputs based on content and light effects image. Additionally, we use only the background portion of the light effect to avoid affecting the foreground person.
}

\textcolor{black}{\textbf{Training stage 2.}}
In the stage 2, we inherit and fix the LoRA weights trained in the stage 1, and train an additional ControlNet~\citep{controlnet} to inject the light effects image into the generation process. To ensure the effectiveness of the ControlNet training, only the content image is used as input during this stage.

\textcolor{black}{\textbf{Inference stage.}}
The light extraction model first extracts light effects from a reference image, which are then fed into the ControlNet.
Concurrently, the background light effect is directly added to the content image, \textcolor{black}{providing a more direct and effective condition control.} 
After encoding into latent space via the VAE encoder~\citep{vae}, the result is concatenated channel-wise with Gaussian noise and fed into the generator, yielding the final output through multi-step denoising.
\textcolor{black}{Notably, our \name{} allows for flexible adjustment of the following light effect attributes on the target image:}
\begin{itemize}
    \item \textcolor{black}{\textbf{Position.} Apply horizontal or vertical translation to the light effect image.}
    \item \textcolor{black}{\textbf{Direction.} Perform flipping or rotating operations on the light effect image.}
    \item \textcolor{black}{\textbf{Intensity.} Modify the multiplicative coefficient of the light effect image overlaid on the content image.}
\end{itemize}

\begin{figure*}[t]
\centering
\includegraphics[width=1\linewidth]{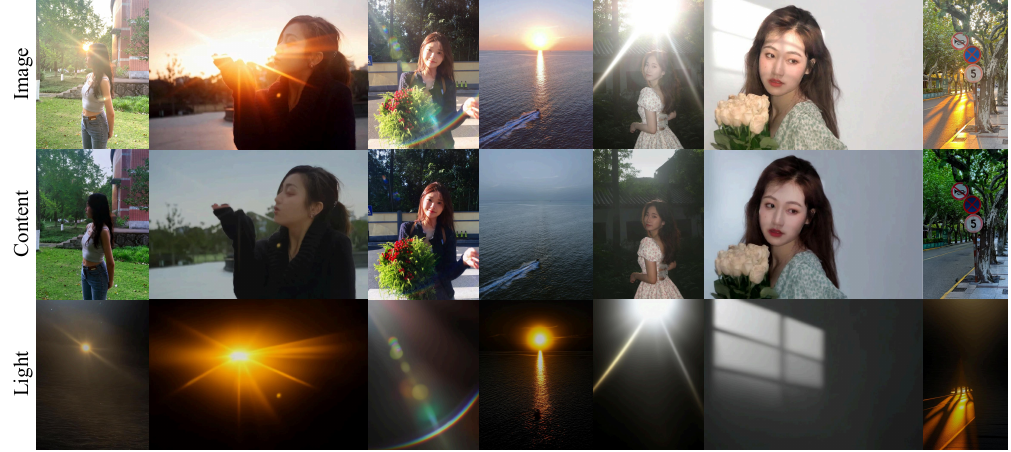}
\caption{\textbf{Visualizations of our generative decoupling. }Our light removal model can eliminate light effects from images while preserving the underlying content unchanged as shown in the second line. Additionally, our light extraction model is capable of isolating the light effects from the image without introducing other extraneous objects.}
\label{fig:generative_decoupling_vis}
\end{figure*}

\begin{figure*}[ht]
\centering
\includegraphics[width=0.9\linewidth]{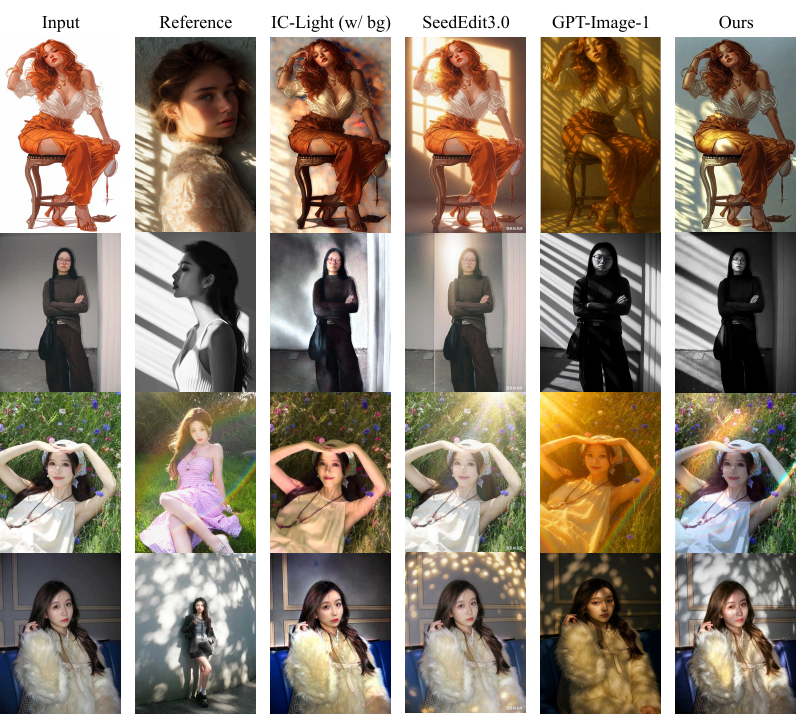}
\caption{\textbf{Visualizations of our \name{}.} IC-Light often exhibit an overall darker tone and may introduce content from the reference image into the resulting images. SeedEdit3.0 can not perceive the type, direction and structure of light effects in the reference image. GPT-Image-1 partially understands the light effects in the reference image, but alters the human ID.}
\label{fig:\name{}_vis}
\end{figure*}

\section{Experiments}
\subsection{Training Details}
\textbf{Generative Decoupling.} We fine-tune the SD1.5~\citep{sd} version of IC-Light~\citep{ic_light} model. We use the AdamW optimizer~\citep{adamw} with the learning rate of 1e-5. We train our model on 8 $\times$ L40s GPUs with a batch size of 128. When training the light removal model, we use a fixed prompt and only require 1K iterations. In contrast, the light extraction model presents a greater training challenge, necessitating 26K iterations.

\textbf{\name{}.} We update the parameters of LoRA and ControlNet using the Adaw optimizer with a learning rate of 1e-5 for the two stages. Considering that our training data scale is on the order of millions, we set the rank of LoRA to 128 and train it for 10K iterations in stage 1. The training process for stage 2 requires 40K iterations. 
We train our \name{} on 16 $\times$ L40s GPUs with a batch size of 128. 

\subsection{Performance of Generative Decoupling}

Considering that the ground truth is not available in advance during the decoupling generation of images, we resort to manual sampling inspection of the generated results to quantitatively evaluate the performance of our model. 
Specifically, we randomly sample 500 image triplets from the large-scale dataset constructed using the pipeline in Figure~\ref{fig:whole_system} (b), and manually evaluate whether the decoupling generation results meet the subjective expectations to calculate the success rate.
The criterion for subjective expectation is that the content image resulting from the decoupling generation should no longer contain noticeable light effects, whereas the light effects image ought to maintain consistency with the light effects in the original image, devoid of any other objects.
Table~\ref{tab:success_rate} presents the detailed statistical results, where we consider both cases: with and without the filtering step shown in Figure~\ref{fig:whole_system} (b).
The statistical results presented in the table demonstrate the efficacy of the filtering step and the high quality of the triplet data constructed via our data construction pipeline.
To further intuitively demonstrate the performance of our generative decoupling, we present visual results of decoupled generations in Figure~\ref{fig:generative_decoupling_vis}.
Our method achieves high-quality decoupled generation on images with various types of light effects across different scenes.

\begin{table}[]
\centering
\caption{\label{tab:success_rate}\textbf{Success rate in Generative Decoupling.}}
\begin{tabular}{cccc}
\toprule
Filtering? & Content & Light   & Total   \\ \midrule
\ding{53}          & 83.45\% & 66.43\% & 58.74\% \\
\ding{52}        & 96.82\% & 86.14\% & 83.64\% \\ \bottomrule
\end{tabular}
\end{table}

\subsection{Performance of \name{}}

\begin{table}[]
\centering
\caption{\label{tab:psnr}\textbf{Quantitative results.} IC-Light uses its background conditioned model, with ground truth serving as the background input.}
\begin{tabular}{lccc}
\toprule
Method   & PSNR $\uparrow$  & SSIM $\uparrow$   & LPIPS $\downarrow$  \\ \midrule
IC-Light & 15.34 & 0.6784 & 0.2756 \\
Ours     & 19.58 & 0.7931 & 0.1982 \\ \bottomrule
\end{tabular}
\end{table}

In this section, we present the generated results of our method and comparisons with other approaches using both subjective and objective evaluation methods.
For objective comparisons, we first use metrics such as Peak Signal-to-Noise Ratio (PSNR), Structural Similarity Index (SSIM), and Learned Perceptual Image Patch Similarity (LPIPS) to evaluate the similarity between a single generated image and ground truth.
\textcolor{black}{It should be noted that, since representative background replacement relighting methods such as Total Relighting~\citep{total_relighting} and Relightful Harmonization~\citep{relightful} are not publicly available, we adopt the background conditioned model of IC-Light as the comparison baseline.}
We select 1000 triplet samples from the large-scale dataset constructed using the generative decoupling strategy, which are not involved in the training process. 
Our \name{} uses the light effect images from triplets for condition generation. Regarding IC-Light, we apply its background conditioned model, using the ground truth as the background input.

As shown in Table~\ref{tab:psnr}, our method clearly outperforms IC-Light. In addition to the above-mentioned similarity metrics computed on individual images, we believe that calculating the distributional distance between the generated dataset and the ground truth provides a more objective evaluation of the overall model performance. Therefore, we sample 12,000 triplets that are not used during training. We compute the Fr\'echet Inception Distance (FID)~\citep{gans} between the generated datasets and the real image set to evaluate the lighting transfer capability of the model, which we refer to as \textbf{Light FID}. We report the result in Table~\ref{tab:light_fid}. When performing inference using the background conditioned model of IC-Light, the ground truth is also directly used as the background input. 
\textcolor{black}{Our method demonstrates significant advantages over IC-Light. Moreover, using only the light effects from the light extraction model and adding them to the content image yields better results than IC-Light.}

% \st{For subjective comparisons,}
\textcolor{black}{For subjective comparison in real-world application scenarios,} we show some visualizations in Figure~\ref{fig:\name{}_vis}.
To fully highlight the advantages of our \name{}, we also show the result of the SOTA relighting method IC-Light, alongside the closed-source SOTA image editing methods SeedEdit3.0~\citep{seededit3.0} and GPT-Image-1\footnotemark[1].
The reference image is used as background input for IC-Light's background conditioned model. For SeedEdit3.0 and GPT-Image-1, we directly provide both input image and reference image to the model simultaneously. We experiment with different prompts to instruct the model about our expected light effects transfer requirements. Subsequently, we select the best output from the results for demonstration purposes.
As illustrated in Figure~\ref{fig:\name{}_vis}, current relighting and image editing methods are unable to fulfill our expected light effect transfer requirements.
\textcolor{black}{IC-Light incorporates the content of the reference image into the generated results, yet the overall style tends to be darker. Although SeedEdit3.0 is capable of modifying the light effects in the target image, it fails to transfer light effects according to user preferences from a reference image. The light effect transfer capability of GPT-Image-1 most closely resembles that of our proposed approach, as it can, to some extent, understand the type and structure of light effects from the reference image. Nevertheless, the generated results show noticeable changes in the character appearance.}
Moreover, free adjustment of light effects position and direction enables more natural integration with the target image, resulting in visually superior outcomes as shown in Figure~\ref{fig:first_page_vis}.
Our \name{} represents the first successful implementation of image-guided customized lighting control via transferring light effects from a reference 
image to a target image.
\footnotetext[1]{\url{https://platform.openai.com/docs/models/gpt-image-1}}

\begin{table}[]
\centering
\caption{\label{tab:light_fid}\textbf{Light FID.} ``Content" in the first row refers to result of lighting-removed content image. ``Content+Light" refers to the result formed by directly adding content and light effect maps.
IC-Light uses its background conditioned model, with ground truth serving as the background input.
}
\begin{tabular}{lc}
\toprule
Method    & Light FID $\downarrow$ \\ \midrule
Content  & 10.61     \\
Content+Light  & 8.40 \\
IC-Light  & 10.05     \\ \midrule
w/o LoRA & 8.44 \\
Synthetic data & 7.72 \\
Unfiltered data &  6.49\\
Ours      & 6.02      \\ \bottomrule
\end{tabular}
\end{table}

\subsection{Ablation Study}
We adopt the Light FID as the objective evaluation metric.

\textbf{Model structure.}  To first verify the effectiveness of the LoRA training in stage 1 of our training pipeline, we conduct an experiment using only stage 2 training. The result is presented in Table~\ref{tab:light_fid}. Without LoRA fine-tuning, the generation results are influenced by the inherent characteristics of IC-Light~\citep{ic_light}, leading to overall color tones that are overly yellow and dark, which in turn causes a significant deviation from real-world data.

\textbf{Training data.} In addition, we conduct an experiment in which \name{} is trained on synthetic data instead. Specifically, we employ the text-conditioned model of IC-Light to relight real images that originally contain no light effects, and then extract the corresponding light effects images from the generated results to form the triplet data. Using this approach, we also generate one million training samples. The experimental result is alse reported in Table~\ref{tab:light_fid}. This result further justify the rationality of our motivation to construct triplets for training by leveraging real-world data through generative decoupling.

\textcolor{black}{\textbf{Data filtering.} In previous sections, we validate the effectiveness of the third step, filtering in the receipt construction pipeline through human subjective evaluation. Here, we conduct further experimental verification. Specifically, we train our model using one million unfiltered triplet samples and compute the FID score as shown in Table~\ref{tab:light_fid}.}

\section{Conclusion}

In this paper, we propose \name{}, the first lighting-control generation method that enables the transfer of light effects from reference images to target images. To achieve the lighting transfer task, we introduce generative decoupling strategy, which decouples the content and light effects in real-world images by fine-tuning two diffusion models. Based on this strategy, we construct over one million high-quality triplet samples, which are then used to train our \name{}. Experimental results demonstrate that our method achieves impressive lighting transfer capabilities, enabling flexible customization of lighting-controlled image generation.
Our \name{} provides a solid and flexible foundation for future developments in image-guided light effect transfer, and we are highly optimistic about its promising application potential in practical lighting editing tasks.

\bibliography{aaai2026}

\appendix
\section{Appendix}
\section{Dataset Details}
\subsection{Light Material}
During the training of both the light removal and light extraction models, we use approximately 100K light material images, of which 10K are sourced from public datasets~\citep{flare7k, flare_removal}, and the remaining 90K are generated using the FLUX.1-schnell~\citep{flux}. We employ the following prompt for generation: \textit{Generate a light material image: $<$random-example$>$, the background color is pure black, true and natural, do not show anything other than light effects.} The ``random-example'' refers to a randomly selected prompt from our designed set of diverse descriptions of light effects. We provide some examples below:
\begin{itemize}
    \item Mystical light beams traversing a clear yet particulate-rich environment, showcasing the diffusion and beauty of light.
    \item Organic patterns of random color and random shape light forming soft, diffused spots with varied shapes and sizes, as if passing through a complex, uneven medium.
    \item A single, intense light source creating a bright spot. The light should have a soft glow around it, with subtle gradients transitioning from bright white to deep black. Include lens flares and light streaks for added realism.
    \item Abstract a small light with curved rainbow effect, no objects, vibrant light streaks, dynamic glow, detailed texture, photorealistic.
    \item An abstract underwater scene featuring dynamic light rays piercing through a rippling water surface. The surface is textured with subtle waves and ripples. The overall atmosphere is ethereal, focusing solely on the interplay of light and water.
    \item Abstract warm oval patches of light beams casting soft shadows on a dark background wall, mimicking the dramatic lighting often used in portrait photography. Focus on the interplay of light and shadow without any objects or figures.
    \item The glow reflected on the lake surface, with its floating light leaping like gold.
    \item The golden side light projects regular light and dark patterns on the deep black background, suggesting the warmth and depth of the hidden light source.
\end{itemize}

\begin{figure*}[ht]
\centering
\includegraphics[width=0.95\linewidth]{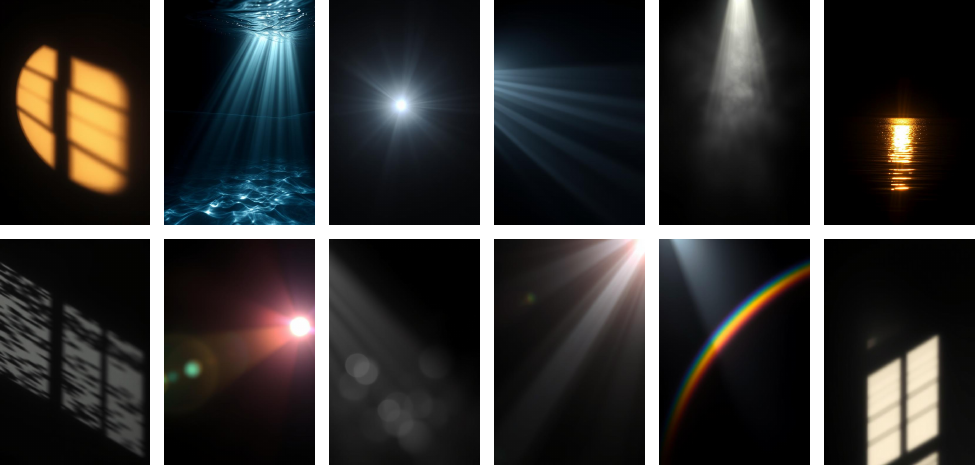}
\caption{\textbf{Visualizations of the light material generated by Flux.}}
\label{fig:Flux_light_vis}
\end{figure*}

We provide some examples of light material generated by Flux in Figure~\ref{fig:Flux_light_vis}.

\subsection{Image-content-light Triplets}
We first employ the vision-language model InternVL2.5~\citep{internvl} to filter approximately 2 million images with prominent light effects from a collection of over 20 million natural images.
Then, we apply our generative decoupling strategy to these datasets for decoupled generation. After the third filtering operation, approximately 1.2 million high-quality triplets that meet expectations are obtained. To accommodate the training requirements of the diffusion model, we use InternVL2.5 to generate brief captions for the images.

\section{Experiment Details}
\subsection{Light Removal Model Training}
The objective of training the light removal model is to enable the model to remove prominent light effects from natural images and generate corresponding content-only images. During the training process, the input is divided into two parts: 80\% consists of random combinations of no light images and light material images, while, to enhance the model's generalization capability significantly, the remaining 20\% is set as the relighting results of no light images. We employ the text-conditioned model from IC-Light~\citep{ic_light} to relight the no light images, without applying background masking to the input images.

\subsection{Light Extraction Model Training}
The objective of training the light extraction model is to enable the model to extract prominent light effects from natural images and generate corresponding light effect images. The input is the random combination of no light images and light material during the training process. In contrast to the training procedure of the light removal model, we perform object masking when overlaying light material images onto no light images, treating them as background light. This approach not only enhances the model’s generalization capability but also mitigates the risk of preserving structural details of the reference image during the inference process.

\section{More Visualizations}

\begin{figure*}[t]
\centering
\includegraphics[width=1\linewidth]{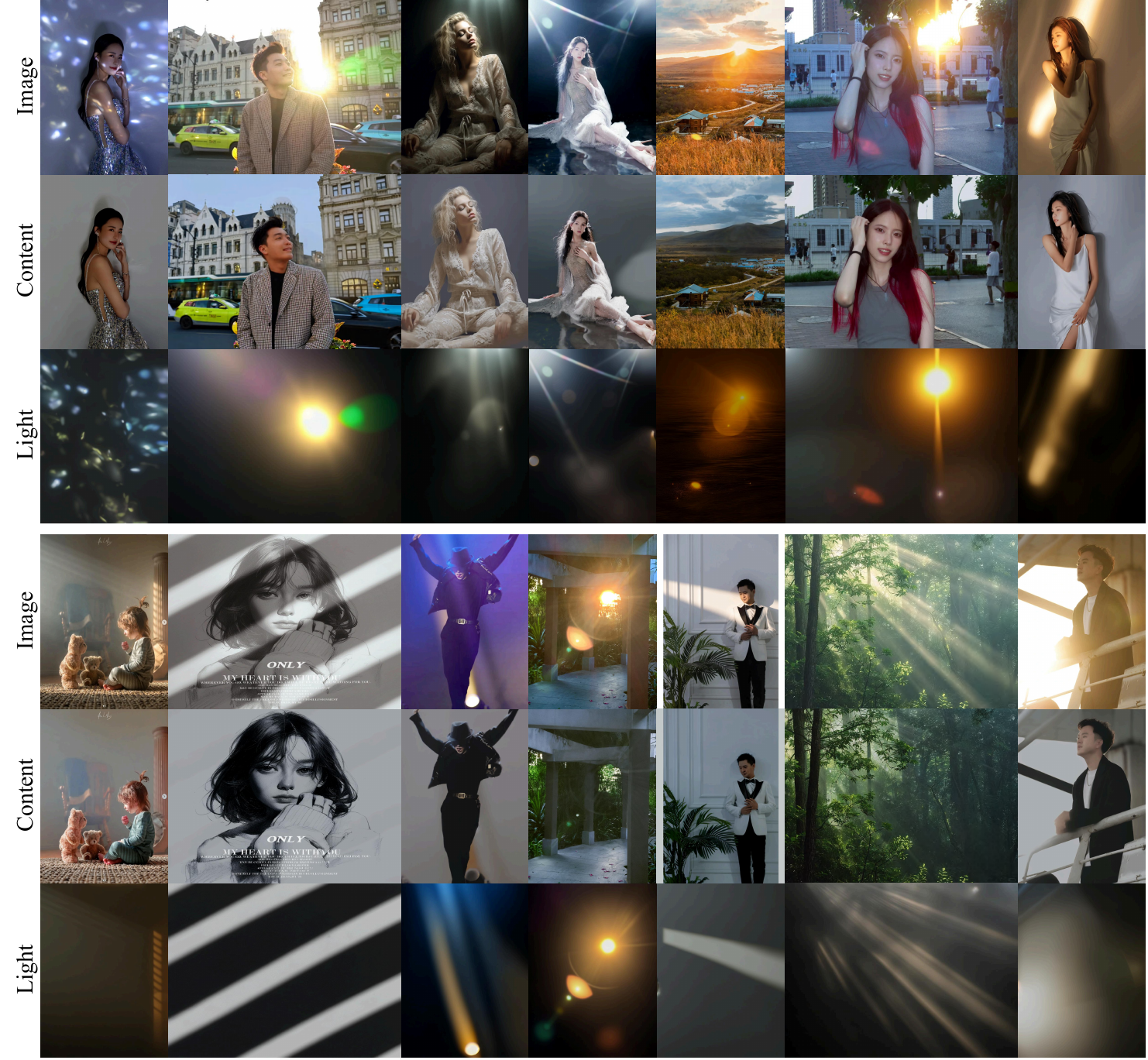}
\caption{\textbf{Visualizations of our generative decoupling.} Our light removal model can eliminate light effects from images while preserving the underlying content unchanged as shown in the second line. Additionally, our light extraction model is capable of isolating the light effects from the image without introducing other extraneous objects.}
\label{fig:generative_decoupling_vis_sup}
\end{figure*}

\begin{figure*}[ht]
\centering
\includegraphics[width=0.9\linewidth]{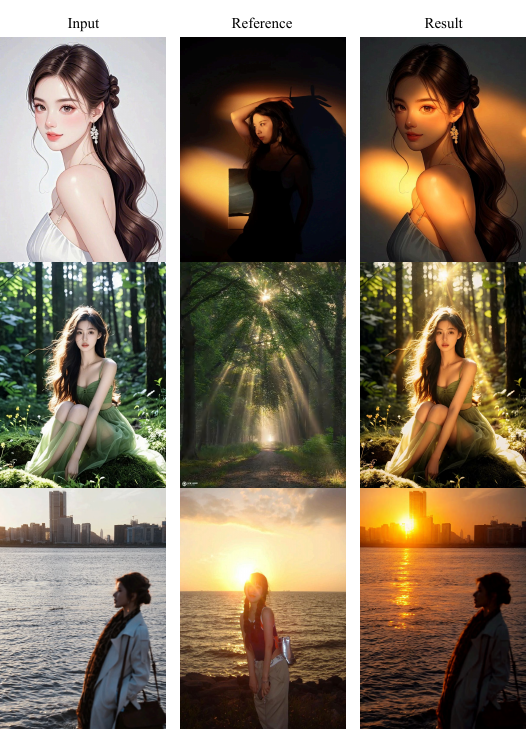}
\caption{\textbf{Visualizations of our \name{}.}}
\label{fig:TransLight_vis_sup}
\end{figure*}

\begin{figure*}[ht]
\centering
\includegraphics[width=0.9\linewidth]{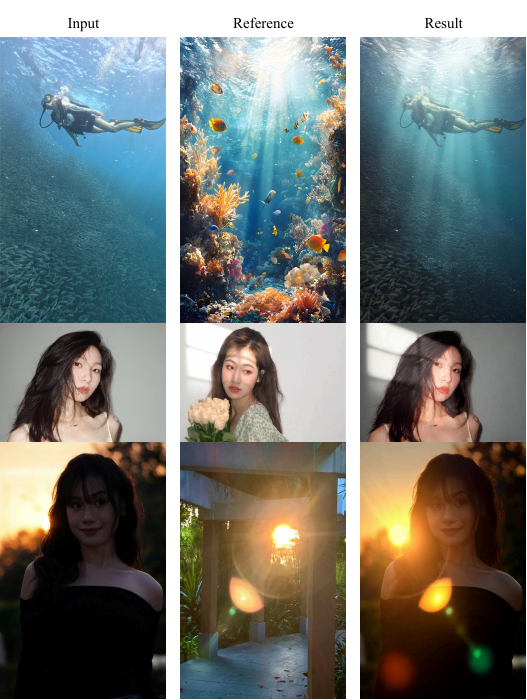}
\caption{\textbf{Visualizations of our \name{}.}}
\label{fig:TransLight_vis_sup_1}
\end{figure*}

\begin{figure*}[ht]
\centering
\includegraphics[width=0.9\linewidth]{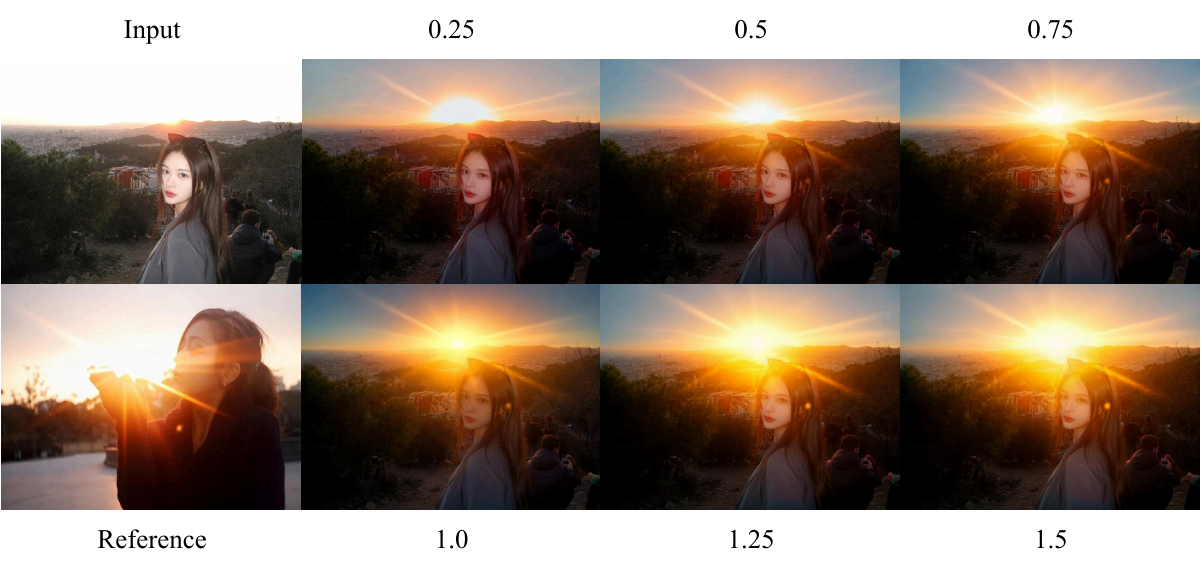}
\caption{\textbf{Transferred results under different intensity coefficients}}
\label{fig:Adjustment_vis}
\end{figure*}

We present more visualizations of our generative decoupling in Figure~\ref{fig:generative_decoupling_vis_sup}. In Figure~\ref{fig:TransLight_vis_sup} and Figure~\ref{fig:TransLight_vis_sup_1}, we show more qualitative results of our \name{} on the task of light effect transfer. In addition to controlling the position and direction of light effects, our \name{} further enables adjustment of their intensity in the transferred results. We show an example in Figure~\ref{fig:Adjustment_vis}.

\section{Discussion}

% \subsection{Limitation}
In this paper, we define a novel task that has not yet received significant attention: light effect transfer. This task aims to transfer the light effects from a reference image onto a user-specified target image. We think such a capability has substantial potential in real-world image-editing scenarios. For instance, after capturing a portrait, users may wish to enhance the visual appeal through appropriate illumination editing. Notably, users typically desire to preserve the original image content while incorporating artistic light effects, which may be sourced from high-quality image templates found on the internet or social media platforms. Motivated by this practical need, we propose TransLight.

Our approach presents a feasible solution to the light effect transfer task by first decoupling the light effect from the reference image into a standalone light-only image, and then compositing it onto the target image. This strategy significantly mitigates the risk of content leakage, and by adjusting the position and orientation of the extracted light image, enables highly flexible and controllable light transfer with great ease. To date, the primary limitation of our TransLight framework lies in its dependence on the performance of the light extraction model. The quality of the final transferred output is heavily influenced by how effectively the light effect is extracted from the reference image. This implies that when the light effect in the user-provided reference is subtle or indistinct, our method may fail to produce satisfactory results. Nevertheless, this does not diminish the practical value of our approach. We have extensively validated our method across diverse reference image scenarios, demonstrating that it achieves lighting editing outcomes that are beyond the reach of existing illumination or image editing techniques. We believe that light effect transfer possesses a broad spectrum of application scenarios and represents a task eminently worthy of in-depth exploration and research. Furthermore, we are confident that our TransLight framework offers a plausible and valuable direction for advancing this field of study.

\end{document}